\documentclass[conference]{IEEEtran}
\IEEEoverridecommandlockouts
\usepackage{cite}
\usepackage{amsmath,amssymb,amsfonts}
\usepackage{algorithmic}
\usepackage{graphicx}
\usepackage{textcomp}
\usepackage{booktabs}
\usepackage{url}
\usepackage{bbm}
\usepackage{graphicx}
\usepackage{amsfonts}
\usepackage{upgreek}
\usepackage{hyperref}
\usepackage{multirow}
\usepackage[table,xcdraw] {xcolor}
\def\BibTeX{{\rm B\kern-.05em{\sc i\kern-.025em b}\kern-.08em
    T\kern-.1667em\lower.7ex\hbox{E}\kern-.125emX}}
\begin{document}

\title{Accurate Segmentation of Optic Disc and Cup from Multiple Pseudo-labels by Noise-aware Learning \\
\thanks{*Corresponding authors (Yang Shen and Shuai Wang)}
}
\author{\IEEEauthorblockN{Tengjin Weng}
\IEEEauthorblockA{\textit{School of Computer Science and Technology}\\
\textit{Zhejiang Sci-Tech University}\\
Hangzhou 310018, China \\
wtjdsb@gmail.com}

\and
\IEEEauthorblockN{Yang Shen\textsuperscript{*}}
\IEEEauthorblockA{\textit{School of Engineering} \\
\textit{Lishui University}\\
Lishui 323000, China \\
tlsheny@163.com}

\and 
\IEEEauthorblockN{Zhidong Zhao}
\IEEEauthorblockA{\textit{School of Cyberspace} \\
\textit{Hangzhou Dianzi University}\\
Hangzhou 310018, China \\
zhaozd@hdu.edu.cn}
\and 

\IEEEauthorblockN{Zhiming Cheng}
\IEEEauthorblockA{\textit{School of Automation} \\
\textit{Hangzhou Dianzi University}\\
Hangzhou 310018, China \\
chengzhiming1118@gmail.com}
\and

\IEEEauthorblockN{Shuai Wang\textsuperscript{*}}
\IEEEauthorblockA{\textit{School of Cyberspace, Hangzhou Dianzi University} \\
Hangzhou 310018, China \\
\textit{Suzhou Research Institute of Shandong University}\\
Suzhou 215123, China\\
shuaiwang.tai@gmail.com}
\and 
}

\maketitle

\begin{abstract}
Optic disc and cup segmentation plays a crucial role in automating the screening and diagnosis of optic glaucoma. While data-driven convolutional neural networks (CNNs) show promise in this area, the inherent ambiguity of segmenting objects and background boundaries in the task of optic disc and cup segmentation leads to noisy annotations that impact model performance. To address this, we propose an innovative label-denoising method of Multiple Pseudo-labels Noise-aware Network (MPNN) for accurate optic disc and cup segmentation. 
Specifically, the Multiple Pseudo-labels Generation and Guided Denoising (MPGGD) module generates pseudo-labels by multiple different initialization networks trained on true labels, and the pixel-level consensus information extracted from these pseudo-labels guides to differentiate clean pixels from noisy pixels. The training framework of the MPNN is constructed by a teacher-student architecture to learn segmentation from clean pixels and noisy pixels. Particularly, such a framework adeptly leverages (i) reliable and fundamental insight from clean pixels and (ii) the supplementary knowledge within noisy pixels via multiple perturbation-based unsupervised consistency. Compared to other label-denoising methods, comprehensive experimental results on the RIGA dataset demonstrate our method's excellent performance. The code is available at \href{https://github.com/wwwtttjjj/MPNN}{https://github.com/wwwtttjjj/MPNN}.
\end{abstract}

\section{Introduction}
Optic disc and cup segmentation is an important step in the automated screening and diagnosis of optic nerve head abnormalities such as glaucoma. With the rapid development of deep learning, various models based on convolutional neural networks (CNNs) have emerged, showcasing promising performance in medical image segmentation. These methods~\cite{long2015fully,badrinarayanan2017segnet,ronneberger2015u} leverage their inherent capacity to learn intricate patterns and features from extensive labeled data, enabling them to achieve high levels of accuracy in segmenting medical images. However, due to the inherent ambiguity of the true segmentation boundary in optic disc and cup segmentation and the inevitable errors in manual labeling, the annotation often contains noise. Numerous studies~\cite{weng2023learning,luo2020semi} have indicated that label noise can significantly impact the accuracy of models. 

To mitigate the impact of label noise on medical image segmentation, one strategy is to use multiple expert annotations.
Ji~\textit{et al.}~\cite{ji2021learning} introduced MRNet, an innovative technique that capitalizes on the specialized insight of individual raters, treating their expertise as prior knowledge to extract high-level semantic features. 
Self-Calib~\cite{Wu2022LearningSO} attempts to find ground truths by running divergent and convergent models in loops, demonstrating a greater capacity for exploiting information from multiple annotations, which results in a more impartial segmentation model. However, it is worth noting that obtaining segmentation annotations from multiple experts is a challenging task. Therefore, it is necessary to explore other available methods.

Without introducing additional expert labels, a series of label-denoising strategies have been proposed to cope with the challenges posed by noisy labels in segmentation tasks. These strategies typically involve two steps: first, noisy labels are identified using various methods, and then there is the option of discarding them or refining them using pseudo-labels.
Some methods~\cite{han2018co,li2021superpixel,zhang2020robust,shi2021distilling} involve training multiple networks and selecting labels with the minimum loss consensus between network predictions for training. 
Other methods~\cite{weng2023learning,zhang2020characterizing,xu2022anti} use Confident Learning~\cite{northcutt2021confident} modules to find noisy labels and use refinement module for processing. 
However, their research often involves artificially simulated noisy labels and annotations from non-experts. There has not been specific research on annotation noise caused by ambiguous boundaries in the segmentation of optic disc and cup.

In this paper, we propose an innovative label-denoising method of Multiple Pseudo-labels Noise-aware Network (MPNN) for accurate optic disc and cup segmentation. 
To address the issue of label noise, we propose the Multiple Pseudo-labels Generation and Guided Denoising (MPGGD) module. We use multiple networks with different initialization to fit the same training set, and after reaching a certain threshold stop training to obtain multiple pseudo-labels for the training set. Subsequently, the pixels corresponding to the labels that perform consistently on all the pseudo-labels are considered clean pixels, and the rest are considered noisy pixels. 
Furthermore, the proposed MPNN consists of a teacher-student architecture, where the student network learns by (i) minimizing the segmentation loss on the set of clean pixels and (ii) minimizing the consistency loss on the set of noisy pixels in the teacher model with multiple uncertainties.
The teacher network is updated based on the parameters of the student network using the Exponential Moving Average (EMA).
Our contributions are as follows:
\begin{itemize}
\item[$\bullet$] To the best of our knowledge, the proposed MPNN is the first method to improve the segmentation accuracy of the optic disc and cup from the perspective of label denoising.
\item[$\bullet$] We propose the MPGGD module, which separates reliable and unreliable information, provides accurate prior information for the network and avoids the negative impact of label noise on the network.
\end{itemize}

\begin{figure*}[!t]
\centerline{\includegraphics[width=\linewidth]{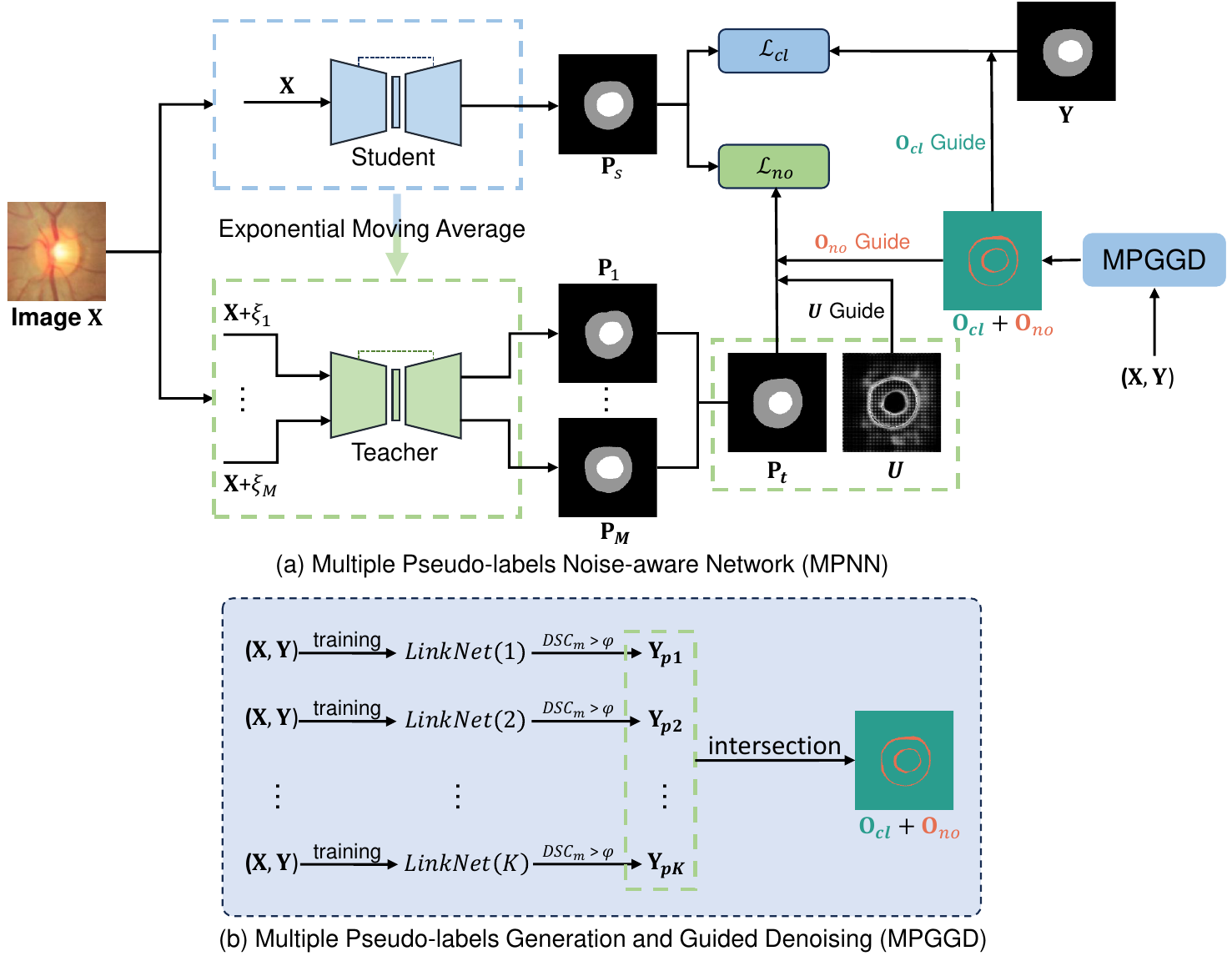}}
    \caption{Illustration of our method. (a) the architecture of MPNN; (b) the MPGGD module. The MPGGD generates multiple pseudo-labels $\{\mathbf{Y}_{p1},...,\mathbf{Y}_{pK}\}$ by multiple different initialization networks ($\{LinkNet(1),..., LinkNet(K)\}$) to fit its $DSC_{m}$ on true label $\mathbf{Y}$ to a certain threshold $\varphi$, which subsequently distinguishes the clean pixels set $\mathbf{O}_{cl}$ from the noisy pixels set $\mathbf{O}_{no}$ by consistent information. The MPNN is constructed by a model consisting of a student network and a teacher network. The student network learns by (i) minimizing the segmentation loss guided by $\mathbf{O}_{cl}$ and (ii) minimizing the consistency loss guided by $\mathbf{O}_{no}$ in the teacher model with multiple uncertainties. The teacher network is updated using the EMA.}
    \label{fig1}
\end{figure*}
\section{Methods}
To simplify the description of our methodology, we define $\mathbf{D}=\{\mathbf X_{(i)}, \mathbf Y_{(i)}\}_{i=1}^{N}$ to represent the dataset, where $N$ represent the total samples. $\mathbf{X}=\{x_1,x_2,...,x_n\}$ denotes that the image $\mathbf{X}$ contains $n = w \times h$ pixels and $ \mathbf{Y} = \{y_1,y_2,...,y_n\}$ indicates the corresponding labels. Here $y \in \{0,1,2\}$ and the categories in $\{0,1,2\}$ denote the background, optic disc, and optic cup respectively. 
Fig.~\ref{fig1} illustrates the method of MPNN. More details will be described in subsequent sections.

\subsection{Multiple Pseudo-labels Generation and Guided Denoising (MPGGD)}
Deep learning models can memorize easy samples and then gradually adapt to more difficult instances as the number of training epochs increases. When noisy labels are present, the deep learning model ends up remembering these incorrect labels, leading to poor generalization performance. 
Therefore, we contemplate halting the training of the network before it fits the noise and generating multiple pseudo-labels corresponding to the training set. These pseudo-labels are utilized to distinguish between clean and noisy pixels.

Specifically, to distinguish clean pixels from noisy pixels, we use multiple networks with different initializations to fit the same training set, and after $DSC_{m}$ (Mean Dice Similarity Coefficient of the optic disc and cup) reaches a certain threshold $\varphi$ stop training to obtain multiple pseudo-labels of the training set. 
To maximize the identification of annotation noise, we adopt an over-correction strategy, that is, pixels corresponding to labels that behave consistently across all pseudo-labels are considered clean pixels, and the rest are considered noisy pixels.

Formally, give an image $\mathbf{X} \in \mathbf{D}$, the ground truth is represented $\mathbf{Y}$. We construct a set of $K$ networks $\{LinkNet(1),..., LinkNet(K)\}$ to generate multiple pseudo-labels $\mathbf{Y}_{p} = \{\mathbf{Y}_{p1},...,\mathbf{Y}_{pK}\}$. All networks are initialized with different values, taking the $DSC_{m}$ as the metric. Considering the scarcity of annotation noise, here $K$ and $\varphi$ are set to 5 and 0.93. 

To differentiate between clean pixels and noisy pixels, we follow the following process:
\begin{equation}
\mathbf{O}_{cl} = \{i \mid \mathbf{Y}_{p1}^i = \mathbf{Y}_{p2}^i =...=\mathbf{Y}_{pK}^i\}_{i=0}^{n}.
\end{equation}

Here, $\mathbf{O}_{cl}$ denotes the separated clean pixels set, and the total number of pixels contained in the set of clean pixels is denoted as $s_{cl} = |\mathbf{O}_{cl}|$. Obviously, the noisy pixels set can be represented as follows:
\begin{equation}
\mathbf{O}_{no} = \{i \mid i \notin\ \mathbf{O}_{cl}\}_{i=0}^{n}.
\end{equation}

The total number of pixels contained in the noisy pixels set is denoted as $s_{no} = |\mathbf{O}_{no}|$. These pieces of information will be used to guide the training of the MPNN.

\subsection{Pixel-level Semi-supervised Pattern Learning}
Inherent ambiguity in segmenting objects and background boundaries in the optic disc and cup segmentation task leads to noisy annotations that impact model performance. Therefore, we study the segmentation task of optic disc and cup with noisy labels and our method draws on the Mean-Teacher~\cite{tarvainen2017mean} model from the classical semi-supervised approach. The student network is parameterized by $\theta$ (updated by back-propagation) and the teacher network is parameterized by $\widetilde{\theta}$ (updated according to the weights $\theta$ at different training stages). The specific computation is the EMA. At training step $t$, $\widetilde{\theta}$ is updated as $\widetilde{\theta}_t = \alpha\widetilde{\theta}_{t-1} + (1-\alpha)\theta_t$, where $\alpha$ is the EMA decay rate, and it is set to 0.99, as recommended by \cite{tarvainen2017mean}.

For a given image $\mathbf{X} \in \mathbf{D}$, we add different perturbations (Gaussian noise $\{\xi_1,...,\xi_M\}$) to $\mathbf{X}$ to generate a set of inputs $\{ \mathbf{X}_m\}_{m=1}^M$. We input $\mathbf{X}$ into the student network to get the corresponding prediction map $\mathbf{P}_s$. At the same time, we input $\{\mathbf{X}_m\}_{m=1}^M$ into the teacher network to get all prediction probability maps $\{\mathbf{P}_m\}_{m=1}^M$. Subsequently, the average of all the prediction maps is taken to get the prediction of the teacher network:
\begin{equation}
\mathbf{P}_t = \frac{1}{M}\sum_{m=1}^{M}{\mathbf{P}_m}.
\end{equation}

The teacher model not only generates goal predictions but also estimates the uncertainty of each goal. Here we follow the approach of~\cite{yu2019uncertainty} and the prediction entropy can be summarized as:
\begin{equation}
\begin{split}
    & u = - \sum_{c}{((\frac{1}{M} \sum_{m=1}^{M}{\mathbf{P}_m^c}})log{(\frac{1}{M} \sum_{m=1}^{M}{\mathbf{P}_m^c}))},
\end{split}
\end{equation}
where $\mathbf{P}_m^c$ is the probability of the $c^{th}$ class in the prediction of $m^{th}$ input. Note that the uncertainty is estimated at the pixel level, and the uncertainty for the entire image $U$ is $\{u\} \in \mathbb{R}^{n}$.
\subsubsection{Clean Pixels Loss}
Clean pixels provide the network with reliable and accurate training signals, thus supporting the networks' ability to learn accurate feature representations and generalization. Our approach extracts a set of clean pixels, enabling the separation of clean and noisy information. Specifically, the clean pixels loss $\mathcal{L}_{cl}$ expressed as:
\begin{equation}
\mathcal{L}_{cl} =\frac{1}{s_{cl}}\sum_{i=0}^{s_{cl}}{\ell_{ce}(\mathbf{P_s}^{\mathbf{O}_{cl}^i},\mathbf{Y}^{\mathbf{O}_{cl}^i}}).
\end{equation}
\begin{table*}[!t]
\centering
    \renewcommand\arraystretch{1.5}
    \caption{Comparative Results of Different Strategies on the Majority Vote Test Set and the Rater1 Test Set, Respectively}
    \label{tab1}
\scalebox{0.9}{
\begin{tabular}{c|cccc||cccc}
\toprule[1pt]
\hline
\multirow{2}{*}{Methods} & \multicolumn{4}{c||}{\textbf{Majority Vote}}                                                                                          & \multicolumn{4}{c}{\textbf{Rater1}}                                                                              \\ \cline{2-4} \cline{5-9}
                         & \multicolumn{1}{c|}{$IoU_{disc}(\%)\uparrow$}                                               & \multicolumn{1}{c|}{$IoU_{cup}(\%)\uparrow$}                                                 & \multicolumn{1}{c|}{$Dice_{disc}(\%)\uparrow$} & $Dice_{cup}(\%)\uparrow$    & \multicolumn{1}{c|}{$IoU_{disc}(\%)\uparrow$}                                               & \multicolumn{1}{c|}{$IoU_{cup}(\%)\uparrow$}                                                 & \multicolumn{1}{c|}{$Dice_{disc}(\%)\uparrow$} & $Dice_{cup}(\%)\uparrow$    \\ \hline \hline
LinkNet~\cite{chaurasia2017linknet}                  & \multicolumn{1}{c|}{83.86} & \multicolumn{1}{c|}{75.77} & \multicolumn{1}{c|}{91.03} & 85.64 & \multicolumn{1}{c|}{81.48} & \multicolumn{1}{c|}{75.66} & \multicolumn{1}{c|}{89.66} & 85.64 \\ \hline \hline
Co-teaching~\cite{han2018co}              & \multicolumn{1}{c|}{84.76} & \multicolumn{1}{c|}{77.25} & \multicolumn{1}{c|}{91.53} & 86.58 & \multicolumn{1}{c|}{82.19} & \multicolumn{1}{c|}{76.30} & \multicolumn{1}{c|}{90.08} & 86.00 \\ \hline
TriNet~\cite{zhang2020robust}                   & \multicolumn{1}{c|}{84.36} & \multicolumn{1}{c|}{77.37} & \multicolumn{1}{c|}{91.29} & 86.71 & \multicolumn{1}{c|}{82.19} & \multicolumn{1}{c|}{76.83} & \multicolumn{1}{c|}{90.10} & 86.36 \\ \hline
2SRnT~\cite{zhang2020characterizing}                    & \multicolumn{1}{c|}{84.50} & \multicolumn{1}{c|}{77.22} & \multicolumn{1}{c|}{91.40} & 86.62 & \multicolumn{1}{c|}{82.64} & \multicolumn{1}{c|}{76.52} & \multicolumn{1}{c|}{90.34} & 86.08 \\ \hline
MTCL~\cite{xu2022anti}                     & \multicolumn{1}{c|}{84.11} & \multicolumn{1}{c|}{76.04} & \multicolumn{1}{c|}{91.15} & 85.83 & \multicolumn{1}{c|}{82.48} & \multicolumn{1}{c|}{74.85} & \multicolumn{1}{c|}{90.26} & 85.05 \\ \hline
PINT~\cite{shi2021distilling}                     & \multicolumn{1}{c|}{84.18} & \multicolumn{1}{c|}{77.13} & \multicolumn{1}{c|}{91.22} & 86.57 & \multicolumn{1}{c|}{82.55} & \multicolumn{1}{c|}{76.11} & \multicolumn{1}{c|}{90.32} & 85.82 \\ \hline \hline
\textbf{MPNN (Ours)}                    & \multicolumn{1}{c|}{\textbf{85.22}} & \multicolumn{1}{c|}{\textbf{78.11}} & \multicolumn{1}{c|}{\textbf{91.83}} & \textbf{87.25} & \multicolumn{1}{c|}{\textbf{83.12}} & \multicolumn{1}{c|}{\textbf{77.34}} & \multicolumn{1}{c|}{\textbf{90.67}} & \textbf{86.77} \\ \hline \bottomrule[1pt]
\end{tabular}}
\end{table*}
\subsubsection{Noisy Pixels Loss}
We choose entropy as a metric for estimating uncertainty. When a pixel label tends to be clean, its predictive probability distribution is likely to be peak, indicating low entropy and low uncertainty. Conversely, if a pixel label tends to be noisy, it is likely to have a flatter probability distribution, indicating high entropy and high uncertainty. Therefore, we consider the uncertainty of each pixel as a pixel noise estimate. Considering the MPGGD method is an over-correction strategy, the set of noisy pixels we separated always contains some easily predictable pixels, and the use of the uncertainty method is to find those clean pixels that can be mistaken as noisy pixels. The consistency loss $\mathcal{L}_{no}$ in the set of noisy pixels expressed as:
\begin{equation}
\mathcal{L}_{no}=\frac{\sum_{i=0}^{s_{no}} \mathbbm{I}({u^{\mathbf{O}_{no}^i} < H})||\mathbf{P}_s^{\mathbf{O}_{no}^i}- \mathbf{P}_t^{\mathbf{O}_{no}^i}||^2}{ {\textstyle \sum_{i=0}^{s_{no}}} \mathbbm{I}({u^{\mathbf{O}_{no}^i} < H})} ,
\end{equation}
where $\mathbbm{I}$ is the indicator function and $H$ (initial value is 0.75 and increases with the number of iterations) is a threshold to select the most certain targets. $u^{\mathbf{O}_{no}^i}$ is the estimated uncertainty $U$ at $\mathbf{O}_{no}^i$ pixel. Note that we only perform the selection of pixels with thresholds higher than $H$ in the noisy pixels set $\mathbf{O}_{no}$.

\begin{figure}[t]
\centerline{\includegraphics[width=\linewidth]{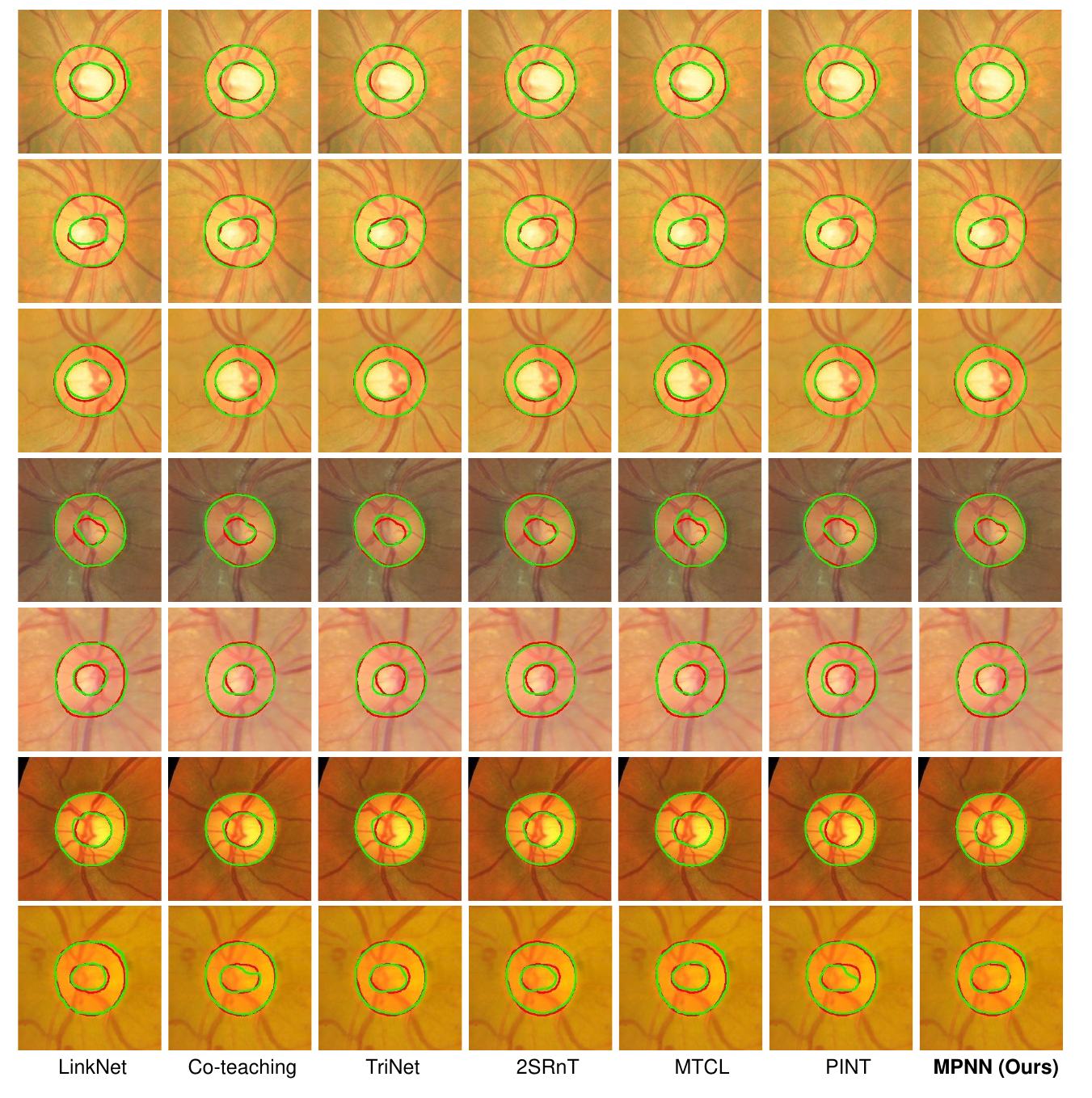}}
    \caption{Visualized segmentation results on the RIGA dataset of different methods. The ground truth mask and predicted mask are represented by red and green, respectively.}
    \label{fig2}
\end{figure}

\subsection{Final Loss}
The total loss function follows the pattern of semi-supervised learning including supervised loss $\mathcal{L}_{cl}$ and unsupervised loss $\mathcal{L}_{no}$. The total loss is expressed as:
\begin{equation}
\mathcal{L}_{total} = \beta \mathcal{L}_{cl} + \lambda \mathcal{L}_{no}.
\end{equation}

Empirically, we set $\beta = 1$ and $\lambda$ is a ramp-up trade-off weight commonly scheduled by the time-dependent Gaussian function \cite{cui2019semi} $\lambda(t) = w_{max} \cdot e^{(-5(1-\frac{t}{t_{max}})^2)}$, where $w_{max}$ is the maximum weight commonly set as 0.1 \cite{yu2019uncertainty} and $t_{max}$ is the maximum training iteration. Such a $\lambda$ weight representation avoids being dominated by misleading targets when starting online training.

\section{Experiments}
\subsection{Dataset and Pre-processing}
The RIGA benchmark~\cite{almazroa2017agreement} is a publicly available retinal disc and cup segmentation dataset consisting of 750 color fundus images from three different sources: 460 images from MESSIDOR, 195 images from BinRushed, and 95 images from Magrabia. Segmentation masks for the disc and cup contours were manually annotated by six glaucoma specialists following the RIGA benchmark~\cite{almazroa2017agreement}. We selected 195 samples from BinRushed and 460 samples from MESSIDOR as the training set as follows~\cite{ji2021learning}. The Magrabia image set containing 95 samples was used as a test set to evaluate the model. We choose the first expert's annotation as the ground truth for the training set. To better evaluate the performance of the method, we construct two test sets, using the masks labeled by the first expert ($\mathbf{Rater1}$) and produced by the majority vote of all experts ($\mathbf{Majority}$ $\mathbf{Vote}$) as ground truth, respectively.

\subsection{Implementation and Evaluation Metrics}
We implemented our method and comparative methods in Python using PyTorch and performed the computations on an NVIDIA GeForce RTX 3090 GPU equipped with 24GB of RAM. We utilized the LinkNet~\cite{chaurasia2017linknet} as the backbone and trained it using the Adam optimizer (betas = (0.9, 0.99)). We set the initial learning rate to 5e-4, decreased it by a factor of 10 every 2000 iterations, conducted a total of 100 training rounds, and maintained a batch size of 8. The input network's image size was scaled to 256 × 256 pixels and normalized using the mean and standard deviation for each channel. To evaluate the model's performance, we calculated the Intersection over Union (IoU) and Dice for each category, excluding the background.


\subsection{Experiments on the RIGA Dataset}
Table~\ref{tab1} presents the quantitative results obtained from various segmentation methods applied to two distinct test sets ($\mathbf{Majority}$ $\mathbf{Vote}$ and $\mathbf{Rater1}$).
Turning our attention to the performance metrics evaluated, the table showcases results for $IoU_{disc}$, $IoU_{cup}$, $Dice_{disc}$, and $Dice_{cup}$, which serve as key indicators of segmentation accuracy. 
We observe that LinkNet while delivering competitive performance, falls short of the MPNN in terms of all metrics. Co-teaching, TriNet, 2SRnT, MTCL, and PINT also exhibit commendable results, but they do not surpass the MPNN outcomes. Among the strategies, it's evident that MPNN emerges as a standout performer. It achieves a remarkable metrics of $IoU_{disc}$ of 85.22\%, $IoU_{cup}$ of 78.11\%, $Dice_{disc}$ of 91.83\%, and $Dice_{cup}$ of 87.25\%. Shifting the focus to the scenario of $\mathbf{Rater1}$, the trends continue to favor MPNN. 
In Fig.~\ref{fig2}, the visualization results of various methods affirm the efficacy of the MPNN. These visualizations demonstrate our method's capacity to precisely segment the targeted regions.
\begin{table}[!t]
\centering
    \renewcommand\arraystretch{1.5}
    \caption{Ablation Study of Our Method on the Majority Vote Test Set. The Impact of Different Numbers of Pseudo-Labels on the Final Results}
    \label{tab2}
\scalebox{0.76}{
\begin{tabular}{c|c|cccc}
\toprule[1pt]
\hline
                          &                     & \multicolumn{4}{c}{\textbf{Majority Vote}}                                                                                                                                                                                            \\ \cline{3-6} 
\multirow{-2}{*}{Method} & \multirow{-2}{*}{$K$} & \multicolumn{1}{c|}{$IoU_{disc}(\%)\uparrow$}                                               & \multicolumn{1}{c|}{$IoU_{cup}(\%)\uparrow$}                                                 & \multicolumn{1}{c|}{$Dice_{disc}(\%)\uparrow$} & $Dice_{cup}(\%)\uparrow$ \\ \hline \hline
                          & 3                   & \multicolumn{1}{c|}{83.52}                                                & \multicolumn{1}{c|}{76.94}                                                & \multicolumn{1}{c|}{90.81}     & 86.41    \\ \ 
                          & 4                   & \multicolumn{1}{c|}{84.90}                                                & \multicolumn{1}{c|}{77.70}                                                & \multicolumn{1}{c|}{91.62}     & 86.97    \\  
                        & 5                   & \multicolumn{1}{c|}{{\color[HTML]{333333} \textbf{85.22}}} & \multicolumn{1}{c|}{{\color[HTML]{333333} \textbf{78.11}}} & \multicolumn{1}{c|}{\textbf{91.83}}     & \textbf{87.25}    \\\multirow{-4}{*}{MPNN}& 6                   & \multicolumn{1}{c|}{{\color[HTML]{333333} {85.19}}} & \multicolumn{1}{c|}{{\color[HTML]{333333} {77.06}}} & \multicolumn{1}{c|}{{91.78}}     & {86.53}\\ \hline \bottomrule[1pt]
        
\end{tabular}}
\end{table}

\begin{table}[!t]
\centering
    \renewcommand\arraystretch{1.5}
    \caption{Ablation Study of Our Method on the Majority Vote Test Set. The Impact of Pseudo-Labels Obtained for Different Threshold Training on the Final Results}
    \label{tab3}
\scalebox{0.75}{
\begin{tabular}{c|c|cccc}
\toprule[1pt] \hline
                          &                     & \multicolumn{4}{c}{\textbf{Majority Vote}}                                                                                                                                                                                            \\ \cline{3-6} 
\multirow{-2}{*}{Method} & \multirow{-2}{*}{$\varphi$} & \multicolumn{1}{c|}{$IoU_{disc}(\%)\uparrow$}                                               & \multicolumn{1}{c|}{$IoU_{cup}(\%)\uparrow$}                                                 & \multicolumn{1}{c|}{$Dice_{disc}(\%)\uparrow$} & $Dice_{cup}(\%)\uparrow$ \\ \hline \hline
                          & 0.91                  & \multicolumn{1}{c|}{84.51} & \multicolumn{1}{c|}{76.97}                                                & \multicolumn{1}{c|}{91.39}     & 86.40    \\ 
                          & 0.92                  & \multicolumn{1}{c|}{84.73}                                                & \multicolumn{1}{c|}{76.99}                                                & \multicolumn{1}{c|}{91.56}     & 86.47   \\
\multirow{-1}{*}{MPNN}    & 0.93                  & \multicolumn{1}{c|}{{\color[HTML]{333333} \textbf{85.22}}} & \multicolumn{1}{c|}{{\color[HTML]{333333} \textbf{78.11}}} & \multicolumn{1}{c|}{\textbf{91.83}}     & \textbf{87.25}    \\
& 0.94                  & \multicolumn{1}{c|}{84.79} & \multicolumn{1}{c|}{ 77.07} & \multicolumn{1}{c|}{91.57}     & 86.51    \\ 
& 0.95                  & \multicolumn{1}{c|}{83.49} & \multicolumn{1}{c|}{ 76.68} & \multicolumn{1}{c|}{90.90}     & 86.25    \\ \hline\bottomrule[1pt]
\end{tabular}}
\end{table}

\subsection{Parameter Ablation Study}  
To determine suitable parameters for the MPNN on the RIGA dataset, particularly during the pseudo-labels generation process, we conducted ablation experiments in the scenario of $\mathbf{Majority}$ $\mathbf{Vote}$.
    
Table~\ref{tab2} presents the effect of different numbers of pseudo-labels on the segmentation accuracy of the MPNN. 
When considering the metric $IoU_{disc}$, the model's performance progressively improves from 83.52\% at $K=3$ to 84.90\% at $K=4$, and further to a notable 85.22\% at $K=5$. Similarly, the metric $IoU_{cup}$ shows improvement from 76.94\% at $K=3$ to 77.70\% at $K=4$, and then to a substantial 78.11\% at $K=5$.
In addition, the Dice coefficients ($Dice_{disc}$ and $Dice_{cup}$) show similar increases. 
However, when $K=6$, a noteworthy observation emerges: the performance metrics exhibit a slight decline compared to the results obtained at $K=5$. 
This observation underlines the importance of striking a balance when introducing pseudo-labels into the training process. While additional pseudo-labels can aid in capturing more label noise, an excessive quantity might lead to the dilution of valuable prior knowledge. This insight encourages further investigation into the optimal amount of pseudo-labels to integrate, considering the trade-off between information enrichment and the risk of introducing noise.

Table~\ref{tab3} presents the impact of multiple pseudo-labels obtained with different training thresholds $\varphi$ on the segmentation accuracy of MPNN.
Notably, for $IoU_{disc}$, we observe a consistent progression as the threshold value increases. The metric shows an improvement from 84.51\% at $\varphi=0.91$ to 84.73\% at $\varphi=0.92$ and eventually peaks at an impressive 85.22\% for $\varphi=0.93$. However, as $\varphi$ increases to 0.94 and 0.95, $IoU_{disc}$ shows a downward trend. When $\varphi$ increases, the resulting clean pixel set may contain inaccurate information, resulting in reduced model performance.
The results affirm the importance of threshold selection during pseudo-label generation, with the highest threshold value of $\varphi=0.93$ yielding the most favorable outcomes in terms of IoU and Dice scores.
\begin{table}[!t]
\centering
    \renewcommand\arraystretch{1.5}
    \caption{Ablation Study of Our Method on the Majority Vote Test Set. Comparison Results of Using Clean Pixels And Noisy Pixels}
    \label{tab4}
\scalebox{0.74}{
\begin{tabular}{c|c|c|cccc}
\toprule[1pt] \hline
\multirow{2}{*}{Method} & \multirow{2}{*}{$\mathbf{O}_{cl}$} & \multirow{2}{*}{$\mathbf{O}_{no}$} & \multicolumn{4}{c}{\textbf{Majority Vote}}                                                                                           \\ \cline{4-7} 
                         &                     &                     & \multicolumn{1}{c|}{$IoU_{disc}(\%)\uparrow$}                                               & \multicolumn{1}{c|}{$IoU_{cup}(\%)\uparrow$}                                                 & \multicolumn{1}{c|}{$Dice_{disc}(\%)\uparrow$} & $Dice_{cup}(\%)\uparrow$    \\ \hline \hline
\multirow{4}{*}{MPNN}   & $\times$                   & $\times$                   & \multicolumn{1}{c|}{84.28} & \multicolumn{1}{c|}{77.35} & \multicolumn{1}{c|}{91.29} & 86.70 \\
& $\times$                   & $\checkmark$                   & \multicolumn{1}{c|}{85.12} & \multicolumn{1}{c|}{77.73} & \multicolumn{1}{c|}{91.75} & 86.94  \\
& $\checkmark$                   & $\times$                   & \multicolumn{1}{c|}{84.97} & \multicolumn{1}{c|}{77.40} & \multicolumn{1}{c|}{91.66} & 86.74 \\
                         & $\checkmark$                   & $\checkmark$                   & \multicolumn{1}{c|}{\textbf{85.22}}  & \multicolumn{1}{c|}{\textbf{78.11}} & \multicolumn{1}{c|}{\textbf{91.83}} & \textbf{87.25} \\ \hline \bottomrule[1pt]
\end{tabular}}
\end{table}
\subsection{Component Ablation Study}
To evaluate the effectiveness of each MPNN component, we performed ablation studies using different variants. The results of our ablation experiments are shown in Table~\ref{tab4}. The evaluation is conducted in the scenario of $\mathbf{Majority}$ $\mathbf{Vote}$.
The symbol $\times$ indicates that the corresponding pixel set is not distinguished when calculating the corresponding loss.
 
\begin{figure}[!t]
\centerline{\includegraphics[width=\linewidth]{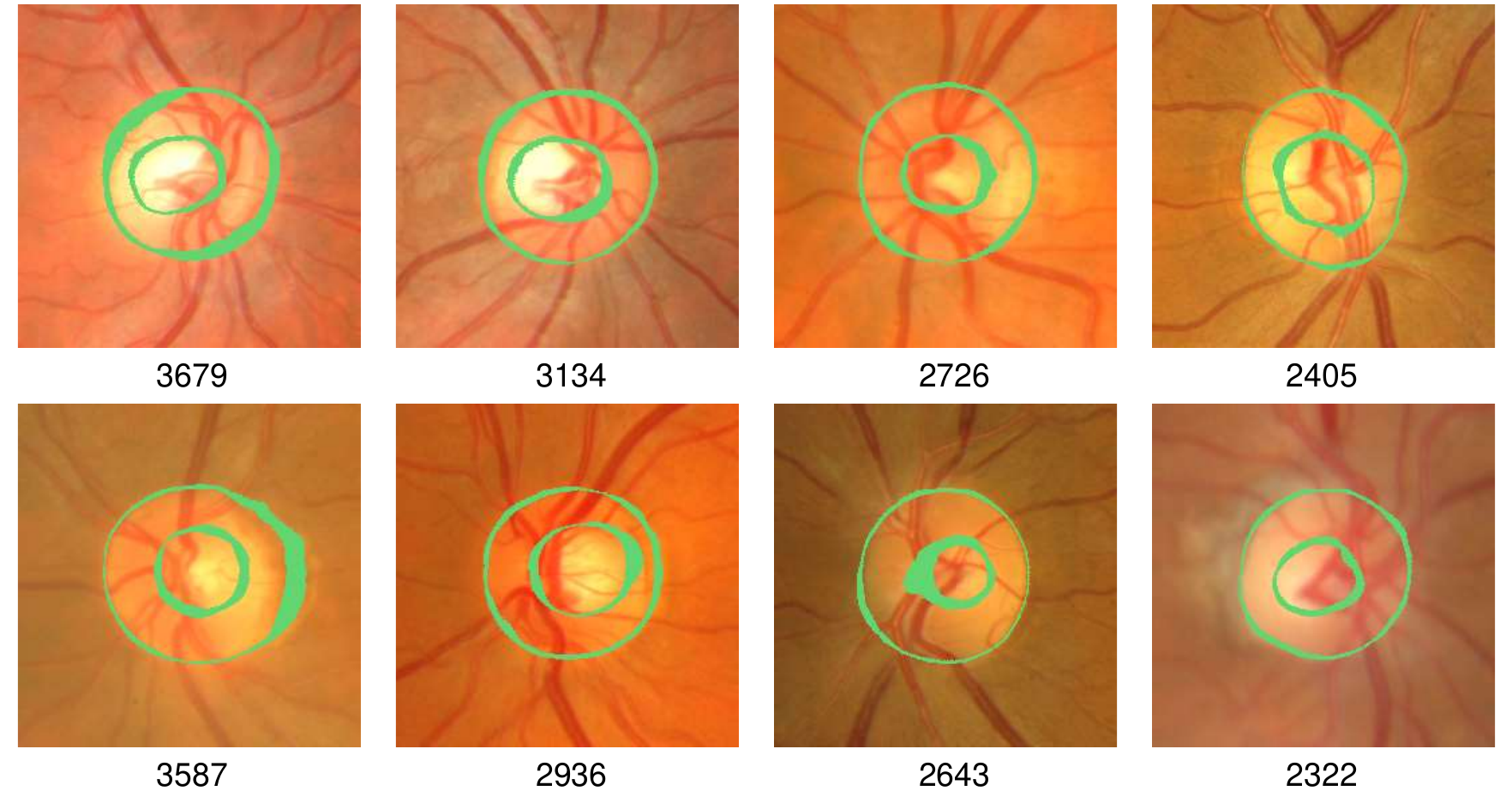}}
    \caption{Visualization of the identified noisy pixels (green), the number underneath indicates the total number of noisy pixels included.}
    \label{fig3}
\end{figure}

When we input noisy pixels and clean pixels without distinguishing between them, the performance of the MPNN is mediocre. As anticipated, when considering only clean pixels or noisy pixels, the MPNN method has demonstrated significant improvement. 
Furthermore, when both clean pixels and noisy pixels are utilized simultaneously, the MPNN method further demonstrates enhanced results. 
Fig.~\ref{fig3} shows our isolated collection of noisy pixels, mostly at the segmentation edges. Our strategy for selecting multiple pseudo-labels is a form of the ``overcorrection-is-necessary” strategy. The filtered noisy pixels necessarily contain instances that the model can confidently predict. This also explains why introducing additional noisy labels for consistency loss yields better results.
\section{Conclusions}
In this paper, we introduce a label-denoising method of the Multiple Pseudo-labels Noise-aware Network (MPNN) for precise optic disc and cup segmentation. 
The proposed MPGGD module uses multiple networks with different initializations to fit the same training set and stops training after reaching a certain threshold to obtain multiple pseudo-labels of the training set. Subsequently, our method can distinguish between noisy and clean pixels by leveraging consensus among multiple pseudo-labels and it proposes distinct learning strategies for noisy and clean pixels based on teacher-student architecture.
Through extensive experimentation on the RIGA dataset, our method performs well on the optic disc and cup segmentation tasks.
The visualized outcomes further validate the exceptional performance and robust denoising capabilities of our proposed method. 
However, there are significant differences among various medical image segmentation tasks with ambiguous boundaries, and such variations can noticeably impact algorithm performance. Therefore, the successful extension of the proposed MPNN to other domains of medical image segmentation with ambiguous boundaries is a direction worthy of in-depth exploration.
\section{Acknowledge}
This research was supported by the Zhejiang Provincial Natural Science Foundation of China under Grant No. LDT23F01015F01 and No. LY21F020004, Natural Science Foundation of Jiangsu Province under Grant No. BK20220266, and National Natural Science Foundation of China No. 61572243 and Construction Project of Lishui University Discipline (Zhejiang Province First Class Discipline, Discipline name: intelligent science and technology) No. XK0430403005.

\bibliographystyle{IEEEtran}
\bibliography{IEEEabrv,myrefs}

\end{document}